\documentclass{llncs}
\usepackage{cite}
\usepackage{amsmath,amssymb,amsfonts}
\usepackage{algorithmic}
\usepackage{graphicx}
\usepackage{textcomp}
\usepackage{xcolor}
\usepackage{listings}
\lstdefinestyle{prompt}{
  basicstyle=\ttfamily\footnotesize,
  breaklines=true,
  breakatwhitespace=true,
  columns=flexible,
  keepspaces=true,
  xleftmargin=0.5em,
  xrightmargin=0.5em,
  frame=single,
  framesep=3pt, % REVISION: 6pt -> 3pt for page limit
  aboveskip=2pt, % REVISION: 4pt -> 2pt
  belowskip=2pt, % REVISION: 4pt -> 2pt
}

\usepackage{latexsym}
\usepackage{mathptmx}
\usepackage[T1]{fontenc}
\usepackage{microtype} % REVISION: standard camera-ready package; recovers lines via protrusion/expansion
\microtypesetup{stretch=40,shrink=40,step=5} % REVISION: slightly wider expansion limits

\usepackage{amsbsy}
\usepackage{subfig}
\usepackage{sidecap}
\usepackage[ruled,vlined]{algorithm2e}
\usepackage{wrapfig}
\usepackage{float}
\usepackage{multirow}
\usepackage{booktabs}
\usepackage{comment}
\usepackage{enumitem}

\usepackage[mathscr]{euscript}

\usepackage[pdftex,colorlinks=true,urlcolor=blue,citecolor=black,anchorcolor=black,linkcolor=black]{hyperref}

\newlist{steps}{enumerate}{1} 
\setlist[steps,1]{label=\textit{Step \arabic*}, leftmargin=*}

\graphicspath{{./figures/}}

\begin{document}

\title{Collaborative Spatial Learning with \\ Multi-LLM Agents in Networked Social Experiments}

\author{Hao He\inst{1}  \and Chris J. Kuhlman\inst{2} \and Xinwei Deng\inst{1}}

\institute{Department of Statistics, Virginia Tech, Blacksburg, VA, USA \\
\email{haoh@vt.edu, xdeng@vt.edu} \and Advanced Research Computing, Virginia Tech, Blacksburg, VA, USA \\ \email{ckuhlman@vt.edu}}

\maketitle

\begin{abstract}

% Collective problem solving requires groups to balance exploration
% against exploitation, and the communication network connecting group
% members mediates this balance. Mason and
% Watts~\cite{mason2012collaborative} showed that human groups in
% shorter-path networks outperform those in longer-path networks on a
% two-dimensional search task, contradicting earlier agent-based
% simulations~\cite{lazer2007network}. We test whether this
% network-efficiency effect transfers to collectives of large language
% model (LLM) agents. Sixteen LLM agents (OpenAI's
% gpt-oss-120b~\cite{openai2025gptoss}) play the Mason--Watts task on
% all eight of the original 3-regular graphs. We benchmark them against
% Gaussian-process Bayesian optimization agents (Expected Improvement,
% Upper Confidence Bound), a uniform random baseline, and human data
% from the original Mason--Watts experiment. The effect replicates
% weakly on cumulative payoff for the LLM agents with randomized
% initialization and for the Bayesian baselines, and is undetectable on
% final-round payoff. A single-sentence prompt change that randomizes
% the first guess yields a performance gain roughly three times the
% range of the path-length effect; without it, the default LLM always
% opens at the grid center. The LLM and Bayesian agents all copy
% neighbors more often than humans, but LLM copying is largely
% positional rather than payoff-conditional. Overall, prompt-level
% design choices matter more than communication topology for LLM
% collective performance.

Collective problem solving often requires that group members consider the tradeoff between exploitation of known solutions and exploration for new ones, where information of known solutions can be disseminated among individual members through communication networks. The Mason--Watts experiment (PNAS 2012) showed that human groups in shorter-path networks outperform those in longer-path networks on a two-dimensional search task. In this work, we focus on the investigation of such network-efficiency effects in the setting of a group of large language model (LLM) agents. Specifically, we consider groups of sixteen LLM agents playing the Mason--Watts experiment on the eight Mason--Watts network topologies. Moreover, we develop mechanistic Bayesian optimization agents such that the performance of LLM agents can be compared with both the mechanistic agents and the human experimental data. Our computational experiments indicate that the LLM agents show a significant network-efficiency effect when instructed to randomize their first-round choices, but not under the default initialization. In this experiment, adding a one-sentence first-round randomization instruction improves collective payoff by more than three times the estimated payoff difference across the eight network topologies. Also, the Bayesian optimization agents obtain higher payoffs than the evaluated LLM agents on this spatial search task. We further compare the agents' exploration--exploitation behavior, copying, and spatial diversity.

\keywords{collective search \and multi-agent LLM systems \and network
structure \and exploration--exploitation \and Bayesian optimization}
\end{abstract}

\section{Introduction}
\label{sec:intro}

In many scientific and engineering applications, it is common to consider collaborative learning in solving a complex problem with a group of problem solvers in a connected network. The problem solvers are required to make tradeoffs between exploration of new solutions and exploitation of known solutions. The network through which members (i.e., problem solvers) share information mediates this tradeoff: dense connections propagate discoveries quickly but can homogenize search, while sparse connections preserve local variation at the cost of slower diffusion. The investigation of such collaborative learning and exploration-exploitation tradeoff is of great interest not only for organizational contexts, but also for multi-agent AI systems, where multiple large language model (LLM) agents coordinate on tasks such as collaborative software development~\cite{qian2024chatdev} and interactive social simulation~\cite{park2023generative}. Whether the communication network affects the collective performance of LLM agents has not been tested empirically, and this gap motivates our work.

\begin{figure}[!t]
    \centering
    \includegraphics[width=0.90\linewidth,trim=0in 1.59in 4.0in 1.2in, % REVISION: width reduced for page limit
      clip,scale=0.50]{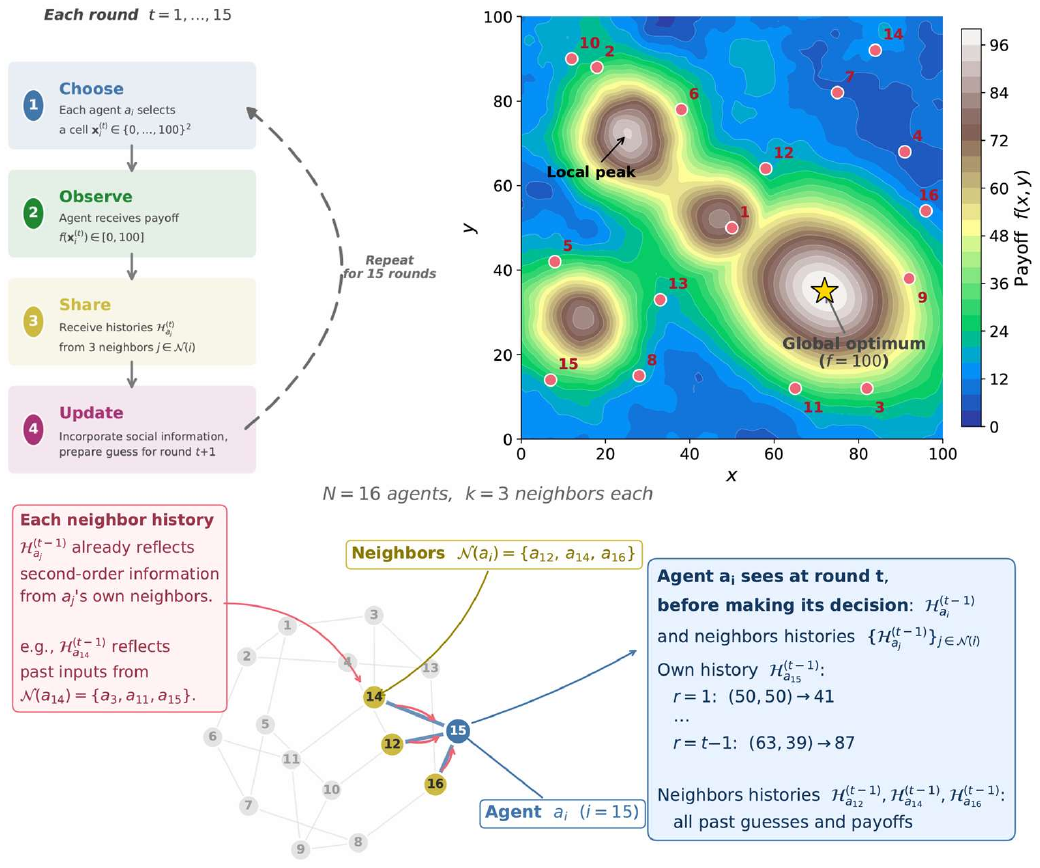}
    \caption{An illustration of the Mason--Watts
experiment~\cite{mason2012collaborative}. The original experiment
placed human participants in these networks. In this work, the
agents are LLM and mechanistic Bayesian agents playing the same
task. \textbf{Upper-left:} each
      round proceeds through four steps---choose a cell, observe its
      payoff, receive neighbor histories, and update---repeated for
      $t=1,\ldots,15$. \textbf{Upper-right:} an example
      $101\times101$ fitness landscape with a unique global optimum
      ($f=100$, gold star) and several local peaks. The red circles represent the (x, y) coordinate selections for the agents in a particular round.
      \textbf{Lower:} each agent~$a_i$ occupies one node of a fixed
      3-regular graph and observes the complete history
      $\mathcal{H}^{(t-1)}_{a_j}$ of each of its three neighbors
      $j\in\mathcal{N}(i)$; each neighbor history itself indirectly reflects
      second-order information from that neighbor's own neighbors.}
    \label{fig:illustration}
\end{figure}

A pioneering empirical benchmark for humans as problem solvers is the Wildcat Wells paradigm proposed by Mason and Watts~\cite{mason2012collaborative}, which we refer to as the Mason--Watts experiment throughout this paper when discussing their study, and as the Wildcat Wells game when describing the task itself. Figure~\ref{fig:illustration} summarizes the paradigm. A group of sixteen networked participants searched a two-dimensional payoff landscape for fifteen rounds, observing at each round the (x, y) coordinates selected and payoff at the coordinates of their three network neighbors, as well as their payoff from their own selected (x, y) coordinates. All participants in
each round make their coordinate choices simultaneously, so a
participant making their choice at round $t$ has only the
information for themselves and their three neighbors for all rounds
up through $t-1$.
Across eight 3-regular graphs spanning a
range of average path lengths, groups in shorter-path networks
consistently earned higher collective payoffs, contradicting an
agent-based simulation by Lazer and
Friedman~\cite{lazer2007network} that predicted the opposite effect.

When considering AI agents to be problem solvers in multi-agent LLM systems, their behavior of exploration-exploitation and sensitivity to the network of collaborative learning has rarely been studied. Conceptually, LLM agents occupy an intermediate position between rule-based agents and human participants: their behavior is neither specified in advance nor acquired through individual experience, but emerges from a pretrained model responding to a structured prompt. Thus, it is important to investigate whether such AI agents reproduce the behavioral regularities that drove the human network-efficiency result, with implications for both multi-agent system design and the theory of network-mediated collaborative learning.

The contribution of this work is fourfold. First, we conduct, to
our knowledge, the first systematic experiment placing groups of LLM agents on varying network topologies to perform sequential
spatial learning, using all eight Mason--Watts graphs with human
data from the original experiment as a behavioral reference.
Second, we construct a reproducible set of $60$ two-dimensional
landscapes at three complexity levels, enabling the investigation
of AI agent behavior across varying landscape difficulty. Third,
we develop Bayesian optimization agents using Gaussian processes to
serve as mechanistic baselines for comparison with LLM search
behavior. These agents incorporate neighbor observations into the
surrogate model, using Expected Improvement~\cite{jones1998efficient}
and Upper Confidence Bound~\cite{srinivas2010gaussian} acquisition
functions adapted for bounded, spatially correlated payoff
surfaces. Fourth, our results yield the following findings. Shorter-path networks yield significantly higher cumulative payoff for the randomized-initialization LLM and both Bayesian agents, but not for the default LLM. By comparison, a single-sentence prompt change at initialization produces a performance gain more than three times the range of the path-length effect. The task-specific Bayesian optimization baselines obtain higher payoffs than the LLM agents. Finally, LLM copying behavior is largely positional rather than payoff-conditional.

The remainder of the paper is organized as follows. Section~\ref{sec:rw}
reviews related work. Section~\ref{sec:design} specifies the task,
agents, and measures. Section~\ref{sec:results} reports the main
findings. Section~\ref{sec:summary} discusses implications,
limitations, and extensions. The data and analysis code are publicly available in the project's GitHub repository.{\renewcommand{\thefootnote}{$\dagger$}\footnote[1]{\url{https://github.com/haohe13/llm-collaborative-spatial-search-data}}}
\section{Related Work}
\label{sec:rw}

Lazer and Friedman~\cite{lazer2007network} used agent-based
simulations on rugged fitness landscapes to argue that less
connected networks preserve exploratory diversity and can
outperform more connected ones on complex problems. Mason and
Watts~\cite{mason2012collaborative} tested this prediction
experimentally and found the opposite: shorter-path networks produced
higher collective payoffs even on complex landscapes. They attributed
the discrepancy to a behavioral feature absent from the simulation:
human participants copied neighbors partially and noisily, preserving
enough diversity that efficient diffusion remained beneficial on
balance. Later studies identified conditions that moderate the network-efficiency effect. Barkoczi and
Galesic~\cite{barkoczi2016social} showed that the relationship
between network efficiency and performance depends on the social
learning strategy agents employ. Bernstein, Shore, and
Lazer~\cite{bernstein2018intermittent} found that intermittent
communication can outperform persistent interaction by restoring
independent exploration. Almaatouq et
al.~\cite{almaatouq2021task} showed that task complexity moderates
group synergy. A recurring theme across these studies is that the
same topology can help or hinder depending on how agents use social
information.

A separate but related line of research considers collectives of
large language models. Multi-agent LLM architectures have been
applied to debate and
reasoning~\cite{du2023improving}, software
development~\cite{qian2024chatdev}, and social
simulation~\cite{park2023generative}, establishing that LLM
collectives can accomplish complex tasks. More directly relevant to
our setting, Papachristou and Yuan~\cite{papachristou2024network}
studied endogenous network formation among LLM agents and found that
the resulting graphs reproduce stylized facts from the human
social-network literature, and Schoenegger et
al.~\cite{schoenegger2024wisdom} showed that LLM ensembles replicate
the wisdom-of-crowds effect in human forecasting. These studies
establish that LLM collectives can reproduce certain social
regularities, but these works do not vary the communication network or test topology effects on tasks involving human network-efficiency effect.

Prior work has also examined LLM behavior in
exploration--exploitation settings. Krishnamurthy et
al.~\cite{krishnamurthy2024explore} found that LLMs do not robustly
explore in multi-armed bandit environments without prompt
interventions such as chain-of-thought reasoning. Harris and
Slivkins~\cite{harris2025explore} evaluated LLMs as separate
exploration and exploitation oracles, and found that they can suggest
useful candidates in large action spaces but underperform simple
baselines on exploitation. Zhang et al.~\cite{zhang2025comparing}
compared LLM and human strategies on bandit tasks and found that
reasoning-enabled LLMs approximate human-like exploration in
stationary settings but fall short under non-stationarity. Findings in these studies are version-dependent, since LLM behavior differs substantially across model generations, which motivates our use of a single fixed open-weight model.

To the best of our knowledge, no prior work has placed groups of LLM agents on varying network topologies to perform sequential
spatial search, nor compared their behavior against mechanistic
Bayesian optimization baselines in such a setting. We do so using
the Mason--Watts paradigm~\cite{mason2012collaborative}, which
provides a well-studied human reference and a controlled framework
for isolating network effects.
\section{Experimental Design and Data Collection}
\label{sec:design}

\subsection{Landscape construction, task and communication networks}
\label{sec:task}

We adopt the Wildcat Wells collective search
paradigm~\cite{mason2012collaborative}. A group of $N = 16$ agents
$\{a_1, \ldots, a_N\}$ searches a discrete two-dimensional grid
$\mathcal{G} = \{0, 1, \ldots, 100\}^2$ over $T = 15$ synchronous
rounds. A hidden fitness landscape
$f : \mathcal{G} \to [0, 100]$ assigns a payoff to each cell. Agents
are connected by a fixed, undirected graph $G = (V, E)$ with
$|V| = 16$ and degree $k = 3$, so that each agent observes three
neighbors. At each round~$t$, every agent~$a_i$ simultaneously
selects a cell $x_i^{(t)} \in \mathcal{G}$ and receives
$f(x_i^{(t)})$. Before its round-$t$ selection, $a_i$ observes its
own history
$\mathcal{H}_i^{(t-1)} = \{(x_i^{(\tau)},
f(x_i^{(\tau)}))\}_{\tau=1}^{t-1}$ and the histories
$\mathcal{H}_j^{(t-1)}$ of each neighbor
$j \in \mathcal{N}(i)$. In this experiment, the goal of each agent
is to maximize its cumulative payoff
$\sum_{t=1}^{T} f(x_i^{(t)})$.

We construct a bank of 60 landscapes (20 per complexity level) with
a three-stage procedure that extends the Perlin-noise construction
of Mason and Watts~\cite{mason2012collaborative,perlin1985image}.
All landscapes satisfy $\max_{x \in \mathcal{G}} f(x) = 100$ with a
unique global maximum. First, a dominant peak is placed at a
uniformly random location $\mu \in \mathcal{G}$ as an elliptical
Gaussian surface scaled so $f(\mu) = 100$. Second, $K$ secondary
peaks are seeded subject to a minimum-separation constraint, each
with peak height drawn uniformly from $[70, 85]$. Third, a fractal
Perlin noise field~\cite{perlin1985image} is added, and the payoff
distribution is calibrated to match the tail targets in
Table~\ref{tab:complexity} with a clipping step that keeps only the
global-peak region above $85$. Landscape generation is deterministic
given a seed that encodes the grid dimension, complexity code, and
replicate index. All experiments use the same landscape instance for
a given (complexity, replicate) pair, so performance differences
reflect agent behavior rather than landscape variation.

\begin{table}[htbp]
\centering
\caption{Landscape complexity parameters. $K$: number of seeded
local maxima. Tail targets specify the fraction of grid cells above
or below a payoff threshold, enforced by percentile calibration.}
\label{tab:complexity}
\begin{tabular*}{\linewidth}{@{\extracolsep{\fill}}lccc@{}}
\hline
Complexity & $K$ & \% cells with payoff $\geq 80$ & \% cells with payoff $\leq 20$ \\
\hline
Low      & 0 & $\approx 20\%$ & $\approx 30\%$ \\
Moderate & 3 & $\approx 10\%$ & $\approx 40\%$ \\
High     & 8 & $\approx 5\%$  & $\approx 50\%$ \\
\hline
\end{tabular*}
\end{table}

\begin{figure}[!t]
\centering
\includegraphics[width=0.95\linewidth]{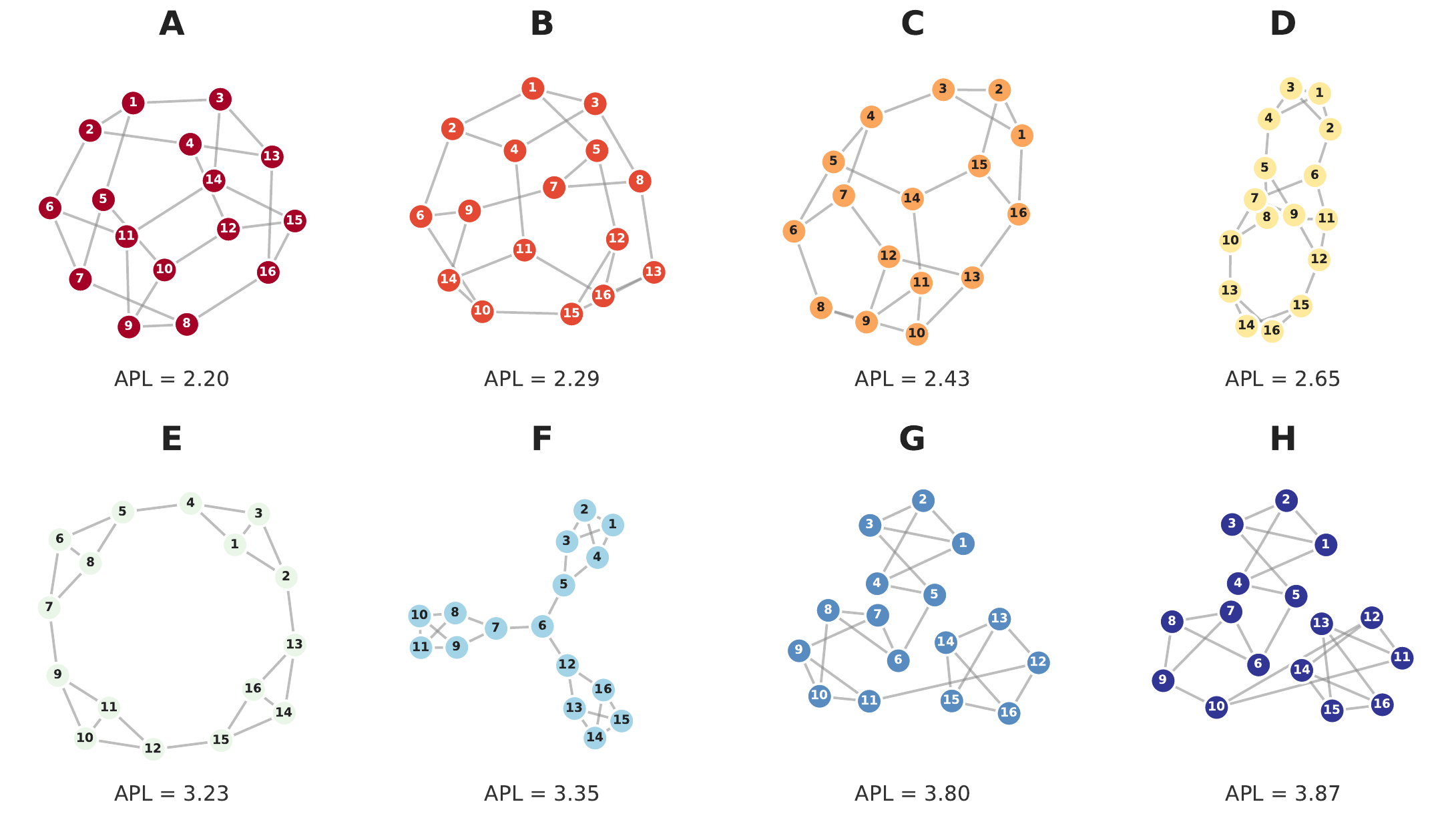} % REVISION: width reduced for page limit
\caption{The eight $3$-regular graphs on $16$ nodes from the
Mason--Watts design~\cite{mason2012collaborative}, ordered by
increasing average path length (APL), which corresponds to
decreasing graph efficiency from A to H. The color scale
(red $\to$ blue) matches the graph-level colors used in
Figure~\ref{fig:graphwise}.}
\label{fig:topologies}
\end{figure}

For communication networks, we use the eight $3$-regular graphs on
$16$ nodes from the original Mason--Watts
design~\cite{mason2012collaborative}, each with $|E| = 24$ edges. Figure~\ref{fig:topologies} displays the eight graphs, labeled A through H in order of increasing average path length (APL).
Following Mason and Watts, we characterize network efficiency by APL:
\begin{equation}
\ell(G) = \frac{1}{n(n-1)} \sum_{u \neq v} d(u, v),
\label{eq:apl}
\end{equation}
where $d(u,v)$ is the shortest-path distance between nodes $u$ and
$v$. Across the eight graphs, $\ell$ ranges from $2.20$ (graph~A) to
$3.87$ (graph~H). Alternative measures such as global
efficiency~\cite{latora2001efficient}, clustering
coefficient~\cite{watts1998collective}, and diameter rank the graphs
similarly; we use APL throughout for consistency with the original
design. We restrict our network choices to these eight graphs for comparability with the human data from the original experiment. Random topologies are left for future work.

\subsection{Agents and experimental design}
\label{sec:agents}

We evaluate four learning agents and a uniform random baseline, and
include a human reference from the original Mason--Watts experiment.
Of the four learning agents, two are LLMs and two are Bayesian
optimization models. For each (complexity, replicate) pair, all five
agent types and all eight graph topologies search the same landscape
instance, so between-agent and between-graph comparisons are not
confounded by landscape sampling. Crossing five agent types with $8$
graphs, $3$ complexities, and $20$ replicates yields
$5 \times 8 \times 3 \times 20 = 2{,}400$ games and $576{,}000$
node-round observations ($115{,}200$ per agent class).

\textit{LLM (default initialization).} Each node is controlled by an
independent instance of OpenAI's
gpt-oss-120b~\cite{openai2025gptoss}, a 117B-parameter
mixture-of-experts model with $5.1$B active parameters per forward
pass, run at the medium reasoning-effort setting. We select this model because its open weights support reproducibility, it follows the strict JSON format reliably, and its mixture-of-experts architecture keeps the cost of the many node-round queries manageable. Each round, an agent receives a structured prompt in two parts. Each round is an independent, stateless model call, so the system message is resupplied every round and the agent's available information is controlled exactly. The system message
specifies the task (a discrete 2D search on an integer grid
$\{0,\ldots,100\}^2$ with payoffs in $[0,100]$ and global maximum
$100$), the total number of rounds ($T = 15$), the agent's role as
one node in a network with access to its own and its neighbors'
histories, the objective of maximizing the sum of its own observed
payoffs, and the required JSON output format. The user message
provides the current round number, the agent's own guess--payoff
history as a list of $(\text{round}, \text{coord}, \text{payoff})$
tuples, and the same history format for each of its three neighbors.
The full prompt template is shown in
Figure~\ref{fig:prompt-template}. The agent returns a JSON object
with a coordinate guess $x_i^{(t)} \in \mathcal{G}$ and a
self-reported belief estimate (expected payoff and uncertainty in
$[0,1]$). Responses are parsed deterministically with no
post-processing of coordinates. The prompt contains no explicit
instruction regarding the first-round guess. The prompt does, however, fully describe the multi-agent context, including the agent's role as one node among networked agents, though it never instructs the agents to coordinate. The center-start in Section~\ref{sec:init} is thus observed behavior given this information, not a consequence of withheld information.

\textit{LLM-RI (random initialization).} This model is identical to
the default LLM except that the round-$1$ system prompt adds a
single sentence instructing the agent to sample its first guess
uniformly from $\mathcal{G}$ (highlighted in
Figure~\ref{fig:prompt-template}). For $t \geq 2$ the prompt is
identical to the default condition. This single modification
isolates the effect of first-round initialization on downstream
collective performance.

\begin{figure}[!t]
\centering

\textbf{System message}
\begin{lstlisting}[style=prompt,basicstyle=\ttfamily\scriptsize]
You are playing a discrete 2-D search game on a grid. Each axis (x, y) uses integer indices in [0,100]. Payoffs are in [0,100]; the global maximum is exactly 100. There are 15 rounds total. On each round, you must choose one grid point to sample.
You are one agent in a network. You can use your own past guesses and payoffs, and the observed history of your neighbors' guesses and payoffs. Your ONLY objective is to MAXIMIZE the SUM of your observed payoffs over all rounds. Higher payoffs are better.
OUTPUT FORMAT (STRICT JSON): respond with EXACTLY one JSON object and nothing else. Schema: {"guess":[<int>,<int>], "belief":{"exp_payoff":<number>,"uncertainty":<number>}}. The value of "uncertainty" MUST be between 0 and 1.
\end{lstlisting}
\textit{LLM-RI only}, appended to the round-1 system message:
\begin{lstlisting}[style=prompt,basicstyle=\ttfamily\scriptsize]
If this is the first round (no observations yet), choose your guess uniformly at random from the grid.
\end{lstlisting}
\textbf{User message (round $t$)}
\begin{lstlisting}[style=prompt,basicstyle=\ttfamily\scriptsize]
You are node {i} in a networked multi-agent game. Grid dimension: 2D, indices: 0..100 for each axis (x, y). Total rounds: 15.
YOUR OWN HISTORY (round, coords, payoff): 1,[x_i^(1)],f(x_i^(1)); ...; t-1,[x_i^(t-1)],f(x_i^(t-1))
NEIGHBOR INFORMATION:
- Neighbor {j_1} history (round, coords, payoff): 1,[x_{j_1}^(1)],f(x_{j_1}^(1)); ...
- Neighbor {j_2} history (round, coords, payoff): ...
- Neighbor {j_3} history (round, coords, payoff): ...
Return ONLY JSON now: {"guess":[...],"belief":{"exp_payoff":...,"uncertainty":...}}
\end{lstlisting}

\caption{Prompt template used for the LLM and LLM-RI agents. Text
in \texttt{\{braces\}} is a placeholder filled at runtime. The
LLM-RI variant differs from the default LLM only in the single
sentence shown in italics, appended to the system message on
round~$1$ (rounds $2 \leq t \leq 15$ use the default system message for
both variants).}
\label{fig:prompt-template}
\end{figure}

\textit{EI and UCB (Bayesian optimization baselines).} Gaussian-process Bayesian optimization is a standard framework for sample-efficient search on smooth, bounded functions~\cite{rasmussen2006gp}, and EI and UCB are standard acquisition rules~\cite{jones1998efficient,srinivas2010gaussian}. We use them as task-specific mechanistic baselines because they receive the same own and neighbor observations as the LLM agents and use explicit exploration--exploitation rules. They provide performance and behavioral references for this spatial task. The EI and UCB agents use the same surrogate model. A Gaussian process with an additive
radial basis kernel
$k(x, x') = k_{\text{short}}(x, x') + k_{\text{long}}(x, x')$
following standard mixture-kernel
practice~\cite{rasmussen2006gp} is fitted to the agent's own observations together with those of its neighbors.
Hyperparameters (length scales $\ell_{\text{short}},
\ell_{\text{long}}$ and observation noise $\sigma^2_\varepsilon$) are
tuned online each round, and the predictive distribution is truncated to respect the bounded payoff
domain. Let $\mu(x)$ and $\sigma(x)$ denote the predictive mean and
standard deviation, and let $f^*$ denote the best payoff observed so
far. The two agents differ only in their acquisition functions:
\begin{gather}
\alpha_{\text{EI}}(x)
  = (\mu(x) - f^*)\,\Phi(z) + \sigma(x)\,\phi(z),
  \quad z = \frac{\mu(x) - f^*}{\sigma(x)},
\label{eq:ei}\\
\alpha_{\text{UCB}}(x) = \mu(x) + \kappa_t\,\sigma(x),
\label{eq:ucb}
\end{gather}
where $\Phi$ and $\phi$ are the standard normal CDF and PDF and
$\kappa_t$ is an adaptive exploration
parameter~\cite{srinivas2010gaussian}. Each agent selects
$x_i^{(t)} = \arg\max_{x \in \mathcal{G}} \alpha(x)$. Once any agent observes the maximum payoff of $100$, no improvement is possible, so EI is near zero everywhere and further UCB exploration only lowers cumulative payoff. That agent's acquisition is therefore frozen, which implements the payoff-maximizing policy of resampling the known optimum. Round-$1$ guesses are drawn from a deterministic
per-agent seed, shared across all eight graphs so that cross-graph
variation reflects only network structure.

\textit{Random (uniform baseline).} On round~$1$, each agent's guess
is drawn from the same per-agent seed used by EI and UCB. On rounds
$2$ through~$15$, each agent draws independently and uniformly from
$\mathcal{G}$, ignoring all neighbor information. This provides a
lower bound that isolates the contribution of learning from
observations.

\textit{Human (reference).} We include data from $29$ sessions of
the original Mason--Watts
experiment~\cite{mason2012collaborative}, in which groups of $16$
participants played the Wildcat Wells game on the same eight
topologies. Each session comprised eight games, one per graph, on
distinct landscapes from the original Mason--Watts bank. Because
these landscapes differ from ours and the original grids are not
publicly available, the human data provide an unmatched reference for the direction of the payoff–topology association and for behavioral measures (copying rate, spatial diversity) that do not depend on landscape identity. We do not compare absolute payoff levels between the human and non-human conditions. Across $232$ games, the overall missingness rate
is $3\%$.

\section{Experimental Results}
\label{sec:results}

The results address four questions: how agents perform across
landscape complexities; whether the Mason--Watts network-efficiency
effect replicates; how initialization shapes default LLM behavior;
and what behavioral patterns explain the observed performance
differences.

\subsection{Performance measures and overview}
\label{sec:perf-overview}

The primary dependent variable is the cumulative game-level mean
payoff,
\begin{equation}
\bar{f}_{\text{cum}} = \frac{1}{NT} \sum_{i=1}^{N} \sum_{t=1}^{T}
f(x_i^{(t)}),
\label{eq:cum}
\end{equation}
where $N = 16$ is the group size and $T = 15$ is the number of
rounds.
This quantity matches the incentive structure and avoids the
ceiling compression that attenuates between-graph variance on
final-round payoff once agents converge spatially by mid-game
(Section~\ref{sec:behavior} below). We also report the final-round
group-mean payoff
$\bar{f}_{T} = (1/N) \sum_{i=1}^{N} f(x_i^{(T)})$ and the near-peak
discovery rate, i.e., the fraction of games in which the group's running
maximum $\max_{i,\,\tau \leq t} f(x_i^{(\tau)})$ reaches at least
$99$ by round~$t$.

\begin{figure}[!t]
\centering
\includegraphics[width=0.84\linewidth]{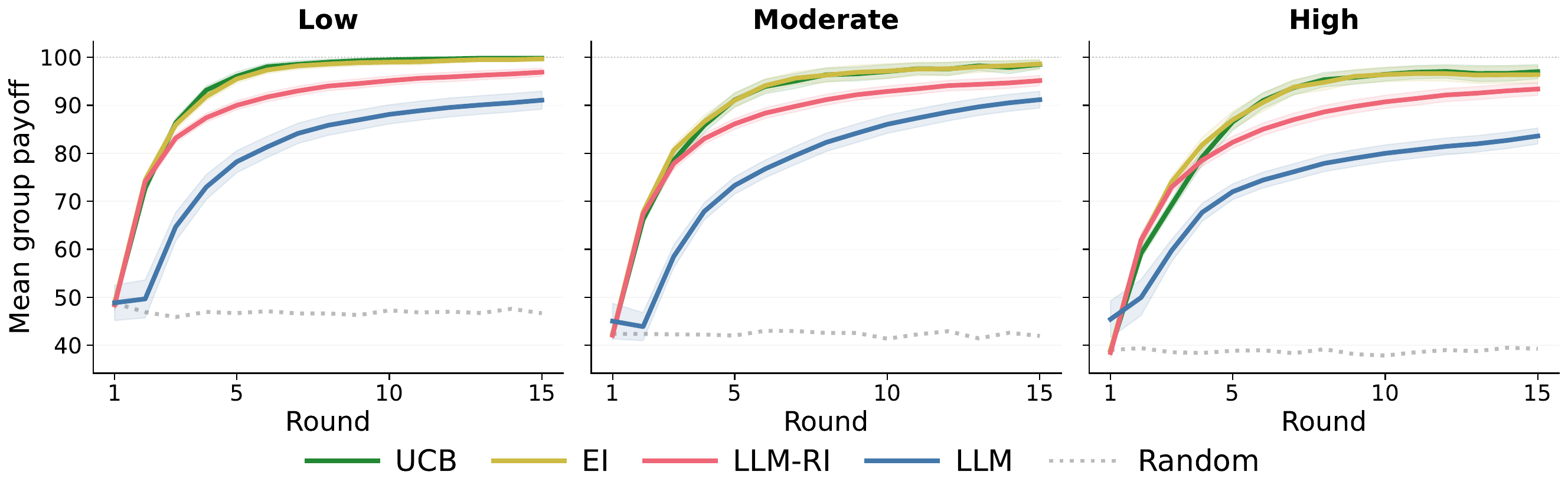} % REVISION: width reduced for page limit
\caption{Group-level learning curves by agent type and landscape
complexity. Each curve plots the round-by-round group-mean payoff
averaged across all eight networks and $20$ replicates per network;
shaded bands are $95\%$ confidence intervals across these games.
The dotted grey line shows the random baseline.}
\label{fig:learning}
\end{figure}

Figure~\ref{fig:learning} plots group-mean payoff over rounds for
the four learning agents and the random baseline, faceted by
landscape complexity. Performance declines from low to high
complexity for every agent, confirming the intended difficulty
gradient. The dominant source of variation is agent type, not
network topology: UCB and EI reach final-round payoffs of
$97$--$100$, LLM-RI attains $93$--$97$, and the default LLM trails
at $84$--$91$. The random baseline is flat at $39$--$47$, confirming
that all other agents learn from the landscape and from each other.
Between-graph differences are small relative to these between-agent
gaps. The two Bayesian optimization agents track each other closely
across all three complexities, while the two LLM variants diverge
substantially; the Bayesian agents separate from both LLM variants
by roughly rounds two through seven, and the gap persists
thereafter.

\begin{figure}[!t]
\centering
\includegraphics[width=0.84\linewidth]{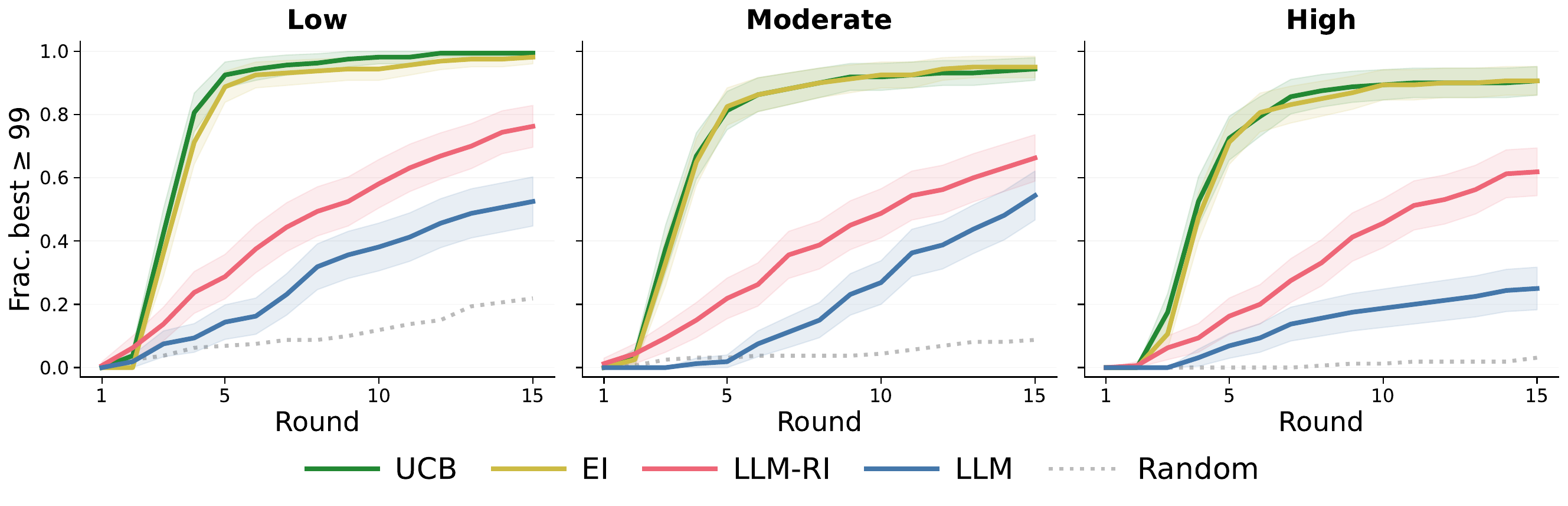} % REVISION: width reduced for page limit
\caption{Cumulative fraction of games in which the group's best
observed payoff reaches at least~$99$, by round. The random baseline (dotted) rarely reaches this threshold.}
\label{fig:peak}
\end{figure}

Figure~\ref{fig:peak} displays the near-peak discovery rates, offering a sharper view of agent performance. By round~$15$, UCB and EI reach a payoff of
at least~$99$ in $91$--$99\%$ of games. LLM-RI succeeds in
$62$--$76\%$, while the default LLM manages only $25$--$54\%$. The random baseline reaches~$99$ in only $54$ of $480$ games, far less often than any learning agent. The default LLM does not merely
earn lower payoffs; it frequently fails to locate the globally
optimal region at all. The comparison from
Figure~\ref{fig:learning} is also visible here: the two Bayesian
models reach by round~$5$ a near-peak discovery rate that the best
LLM variant (LLM-RI) does not attain even by round~$15$.

\subsection{Network structure and collective performance}
\label{sec:network-effect}

Mason and Watts~\cite{mason2012collaborative} reported that groups
in shorter-path networks earned higher cumulative scores. We test
for the same relationship using OLS regressions of game-level mean
payoff on average path length (APL), with complexity fixed effects
for the non-human agents. Cumulative payoff is the primary dependent
variable because it matches the incentive structure and avoids the
between-graph variance compression that affects final-round payoff
once agents converge spatially (Section~\ref{sec:behavior} below).

Because APL takes only eight unique values, the appropriate level of aggregation for inference requires care. Table~\ref{tab:apl} reports two
specifications side by side. The \emph{game-level} specification
uses one row per game ($n = 480$ for each non-human agent,
$n = 232$ for humans), matching the original Mason--Watts analysis.
The \emph{graph-level} specification averages payoffs within each
$(\text{graph}, \text{complexity})$ cell for the non-human agents
($n = 24$) and within each graph for humans ($n = 8$), matching the
level at which APL actually varies. Because the non-human design is
balanced, with exactly $20$ games per $(\text{graph},
\text{complexity})$ cell, the cell means that enter the graph-level
regression are unweighted averages of the game-level observations,
and the point estimate $\beta_{\text{APL}}$ is algebraically
identical at both levels. The two specifications therefore answer
complementary questions with the same slope: the game-level model
tests whether APL predicts individual-game payoff after controlling
for complexity; the graph-level model tests whether APL explains the
residual variation across the eight topologies once within-cell
landscape noise has been averaged away. Reporting both levels shows
whether the coefficient's significance depends on how residual
variance is allocated.

\begin{table}[!t]
\centering
\caption{OLS regressions of cumulative game-level mean payoff on APL
under two specifications. Point estimates of $\beta_{\text{APL}}$ are
identical at both levels because the non-human design is balanced;
the specifications differ only in how residual variance is allocated.
Negative coefficients indicate that more efficient networks yield
higher performance. Non-human models include complexity fixed
effects; the human regression does not. Non-human and human slopes
share sign but not scale (different landscape banks).}
\label{tab:apl}
\begin{tabular*}{\linewidth}{@{\extracolsep{\fill}}lrrrrrrr@{}}
\hline
 & & \multicolumn{3}{c}{Game-level} & \multicolumn{3}{c}{Graph-level} \\
\cmidrule(lr){3-5} \cmidrule(lr){6-8}
Agent   & $\beta_{\text{APL}}$ & SE   & $p$-value      & $n$ & SE   & $p$-value      & $n$ \\
\hline
LLM     & $-0.41$              & 0.82 & 0.619    & 480 & 0.35 & 0.260    & 24 \\
LLM-RI  & $-1.37$              & 0.46 & 0.003    & 480 & 0.39 & 0.003    & 24 \\
UCB     & $-0.99$              & 0.44 & 0.026    & 480 & 0.18 & $<0.001$ & 24 \\
EI      & $-1.26$              & 0.45 & 0.005    & 480 & 0.20 & $<0.001$ & 24 \\
Human   & $-4.65$              & 1.79 & 0.010    & 232 & 2.21 & 0.082    & 8  \\
\hline
\end{tabular*}
\end{table}

For LLM-RI, EI, and UCB, graph-level standard errors are
substantially smaller than game-level ones, with the strongest
difference for EI and UCB ($p$-value $< 0.001$ at the graph level versus
$p$-value = $0.005$ and $p$-value = $0.026$ at the game level). The reason is that
$20$ replicate landscapes contribute to each cell, and their
variance is i.i.d.\ with respect to APL; averaging removes this
noise from the SE denominator, yielding the appropriate precision
for inference about a graph-level property. The default LLM remains
non-significant at both levels.

The human row moves in the opposite direction: game-level
$p$-value = $0.010$, graph-level $p$-value = $0.082$. Here the mechanism differs.
The archival design has one game per $(\text{session},
\text{graph})$ pair, so aggregating from $232$ to $8$ rows removes
genuine across-session variability rather than replicable landscape
noise, and reduces $n$ by a factor of nearly $30$. The human slope
is directionally consistent with the original finding: sign and
magnitude ($\beta = -4.65$) are preserved, but the eight-graph
sample alone does not clear a conventional significance threshold.
None of our conclusions about the non-human agents depend on this
choice. On final-round payoff, all coefficients for the non-human agents
collapse toward zero (all $|\beta| < 0.30$, all $p$-value $> 0.4$),
consistent with the ceiling-compression interpretation developed in
Section~\ref{sec:behavior} below.

Figures~\ref{fig:graphwise} and~\ref{fig:apl-scatter} display the
topology effect in two complementary views.
Figure~\ref{fig:graphwise} plots each graph's baseline-corrected,
normalized deviation from the round-level grand mean. The human
panel shows separation by efficiency rank: efficient graphs (A,~B)
diverge upward, inefficient graphs (G,~H) diverge downward. The UCB
panel shows a qualitatively similar but smaller pattern. The LLM-RI panel shows the same ordering in the early and middle rounds, consistent with its significant slope, while the default LLM panel shows the least systematic separation. Figure~\ref{fig:apl-scatter} re-expresses the
same relationship as a scatter of graph-level means with fitted
regression lines. Lazer and Friedman~\cite{lazer2007network} further predicted that
inefficient networks should outperform efficient ones on rugged
landscapes. We find no support for this: the complexity
$\times$ path-length interaction is non-significant for all four
learning agents (all $F < 1$, all $p$-value $> 0.40$).

\begin{figure}[!t]
\centering
\includegraphics[width=0.83\linewidth]{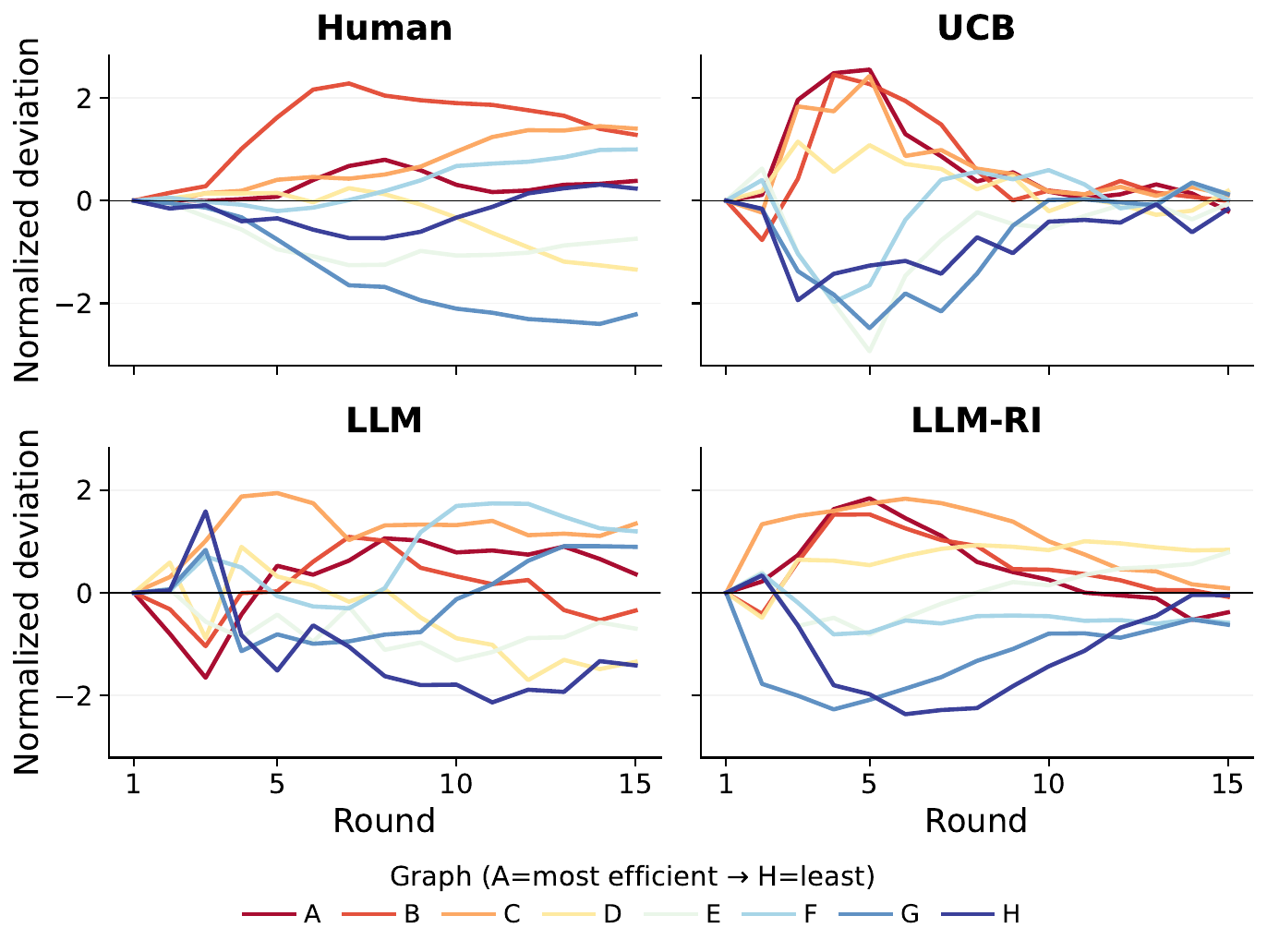} % REVISION: width reduced for page limit
\caption{Baseline-corrected, normalized deviation from the
round-level grand mean, by graph. Colors run from red (A, most
efficient) to blue (H, least efficient). All curves start at zero;
subsequent divergence reflects network-dependent search dynamics.}
\label{fig:graphwise}
\end{figure}

\begin{figure}[!t]
\centering
\includegraphics[width=0.76\linewidth]{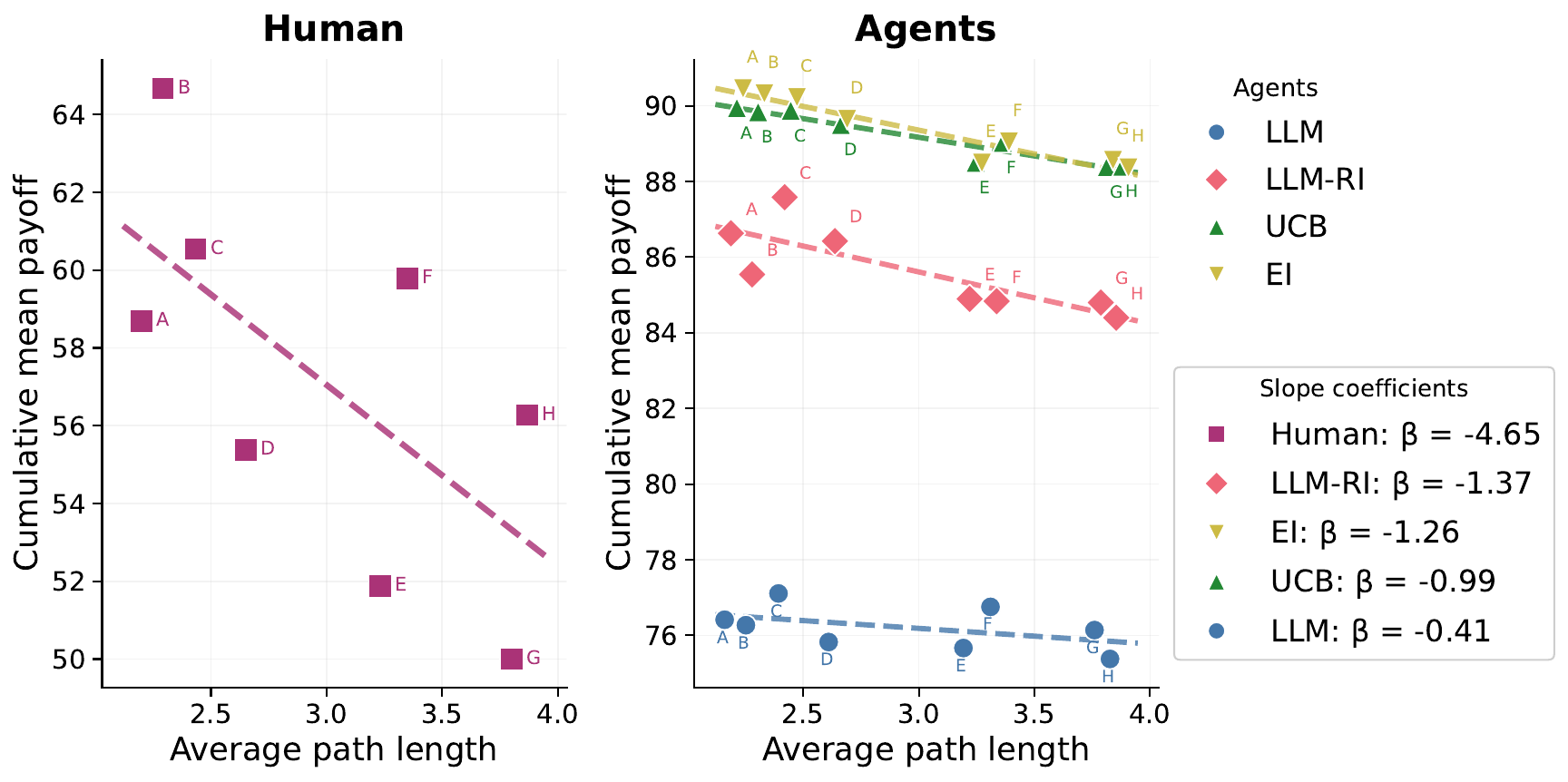} % REVISION: width reduced for page limit
\caption{Cumulative game-level mean payoff versus average path
length. Left: human archival data. Right: non-human agents. Each
point is a graph-level mean; dashed lines are OLS fits matching the
graph-level specification in Table~\ref{tab:apl}.}
\label{fig:apl-scatter}
\end{figure}

\subsection{Initialization effects}
\label{sec:init}

Without an explicit randomization instruction, default LLM agents begin at the grid center $(50, 50)$ in almost all $480$ games ($7{,}677$ of $7{,}680$ first-round choices). We retain this condition as the default prompt without an explicit first-round coordination rule. This
is collectively suboptimal: $16$ agents starting at the same location
lose the spatial coverage that makes early-round neighbor
information valuable. The uniform center-start suggests that the
default model does not spontaneously reason about where the other agents are likely to begin. Recent work has documented that
LLM theory-of-mind performance, measured on tasks requiring
reasoning about other agents' beliefs and actions, is variable
across models and highly sensitive to surface-level prompt
features~\cite{kosinski2024tom,ullman2023large,strachan2024testing}. Our observation is consistent with that pattern. We treat this reading as an interpretation rather than a demonstrated mechanism, since alternatives such as a positional prior toward the grid midpoint cannot be excluded. 

On cumulative game-level mean payoff, LLM-RI exceeds the default
LLM by $9.44$ points (paired $t = 15.83$, $p$-value $< 0.001$, Cohen's
$d = 0.72$); on final-round payoff the gap is $6.53$ points
($t = 10.52$, $p$-value $< 0.001$, $d = 0.48$). For comparison, the entire
range of the path-length effect across the eight graphs is less
than $3$ points for any non-human agent. On final-round payoff,
where all path-length coefficients are non-significant ($p$-value $> 0.5$),
the initialization gap remains highly significant ($p$-value $< 0.001$). In this task setting, the first-round initialization instruction has a larger effect on collective payoff than the choice among the eight topologies. A general ranking of prompt factors against network factors would require a factorial design beyond this study.

The initialization effect is visible throughout the preceding
figures. In Figure~\ref{fig:learning}, the LLM-RI curve lies above
the default LLM across all complexities. In
Figure~\ref{fig:apl-scatter}, the default LLM regression line is
essentially flat ($\beta = -0.41$, $p$-value $= 0.618$), while LLM-RI
recovers a significant slope ($\beta = -1.37$, $p$-value $= 0.003$).

This result clarifies the network findings. The default LLM's
insensitivity to topology partly reflects its homogenized start:
when all $16$ nodes begin at the same point, the first several
rounds are spent dispersing rather than benefiting from neighbor
information. Randomizing the first-round choices recovers a
significant path-length effect. This identifies first-round
initialization as an important design factor in this task.

\subsection{Behavioral mechanisms}
\label{sec:behavior}

We characterize agent behavior with two measures. Copying rate is
the fraction of node-rounds (for $t \geq 2$) in which agent~$a_i$'s
guess falls within Euclidean distance $\varepsilon = 3$ of any
neighbor's previous-round guess:
\begin{equation}
\mathrm{copy}_i^{(t)} = \mathbf{1}\!\left[\,
\min_{j \in \mathcal{N}(i)}
  \| x_i^{(t)} - x_j^{(t-1)} \| \leq \varepsilon \right].
\label{eq:copy}
\end{equation}
The threshold $\varepsilon = 3$ equals $3\%$ of the axis length and tolerates small integer offsets around a copied target.
Spatial diversity is the mean pairwise Euclidean distance among
agents' guesses in round~$t$:
\begin{equation}
D^{(t)} = \frac{2}{N(N-1)} \sum_{i < j}
\| x_i^{(t)} - x_j^{(t)} \|.
\label{eq:div}
\end{equation}
$D^{(t)} \approx 0$ indicates complete convergence; under uniform
random placement on $\{0, \ldots, 100\}^2$,
$\mathbb{E}[D^{(t)}] \approx 52$. These measures describe search
strategy rather than payoff and are therefore valid for cross-source
comparison, including against human data.

\begin{figure}[!t]
\centering
\includegraphics[width=0.86\linewidth]{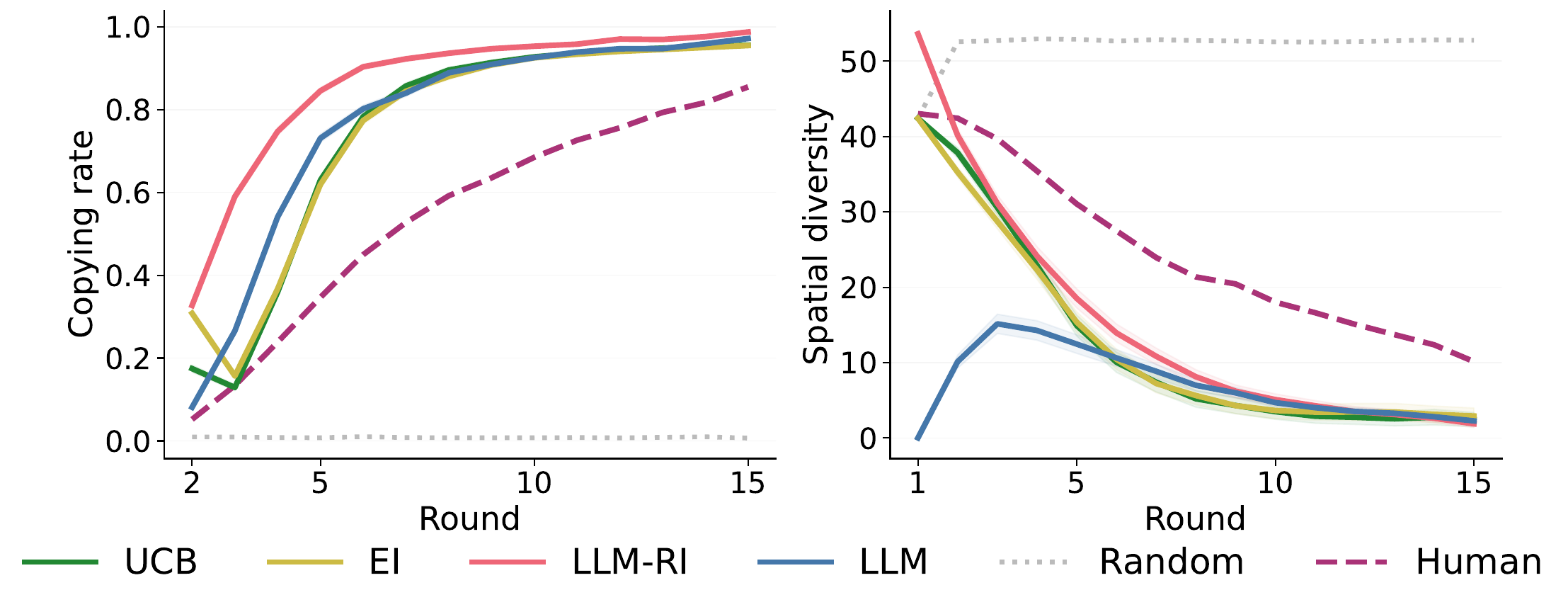} % REVISION: width reduced for page limit
\caption{Copying rate (left) and spatial diversity (right) over
rounds. Human data (dashed) are included as a behavioral reference.
The random baseline (dotted) shows near-zero copying and
near-constant diversity.}
\label{fig:behavior}
\end{figure}

Figure~\ref{fig:behavior} reports copying rate and spatial diversity
over rounds. Human data are included as a behavioral reference, and
the random baseline as a null reference.

All four learning agents copy more than human participants. Mean
copying rates over rounds $2$--$15$ are $86\%$ for LLM-RI, $77\%$
for the default LLM, $75\%$ for EI, $74\%$ for UCB, and $54\%$ for
humans. The random baseline copies at less than $1\%$, confirming
that the metric captures genuine positional imitation. Spatial
diversity collapses correspondingly: mean pairwise distance drops
from $42$--$54$ cells at round~$1$ to $2$--$3$ cells by round~$15$ for EI, UCB, and LLM-RI (the default LLM starts near zero by construction and converges to the same range), versus a reduction from $43$ to
$10$ cells for humans.

These measurements help explain the performance differences we
observed across the network topologies. The rapid collapse in
spatial diversity is one reason why graph topology has
limited influence on final-round outcomes. In the original human
experiments, group-level diversity persisted long enough for
information-propagation speed to
matter~\cite{mason2012collaborative}. In our setting, most nodes
occupy the same region within a few rounds; after that point,
neighbor positional information is largely redundant regardless of
network position. The topology effect that does appear on
cumulative payoff reflects the early rounds, before spatial
homogenization renders network structure irrelevant.

Aggregate copying rates, however, mask a sharp difference in
copying quality. We report the Pearson correlation $r$ between
individual copying rate and individual payoff. For UCB and EI, $r = 0.87$--$0.90$, so copying rate is strongly associated with payoff. For the default LLM, the
same individual-level correlation is $r = 0.25$, and at the
game-level the correlation between collective copying and near-peak
discovery is only $r = 0.27$. LLM-RI is intermediate at the
individual level ($r = 0.42$), and the human reference is at
$r = 0.69$. LLM agents copy more often than any other class, yet their copying is the least associated with payoff, a pattern consistent with 
positional imitation rather than payoff-conditional social learning.
This aligns with the center-start observation: copying frequency is largely unrelated to payoff for the LLM, just as its first guess is unrelated to where the other agents will begin. Together, the common start, rapid diversity collapse, and weak
association between copying and payoff help explain the default
LLM's lower performance relative to that of the Bayesian agents.
\section{Summary and Discussion}
\label{sec:summary}

In this work, we observed that the Mason--Watts
network-efficiency effect transfers only partially to LLM
collectives. On cumulative payoff, shorter-path networks yield
higher performance for LLM-RI and both Bayesian baselines. The direction agrees with the human result, although slope magnitudes are not directly comparable because the landscape banks differ. On final-round payoff,
the network topology effects vanish for all non-human agents. The
default LLM, which starts almost every game at the grid center, shows no
topology effect. All four types of learning agents copy neighbors at
substantially higher rates than humans and end at far lower spatial diversity than the human groups.

The deterministic center-start is consistent with the default
prompt not inducing coordinated first-round behavior, and the low copying--payoff correlation shows that copying is only weakly associated with payoff. These observations align with the reported
prompt sensitivity of theory-of-mind
performance~\cite{kosinski2024tom,ullman2023large,strachan2024testing}
(Section~\ref{sec:init}). In this setting, a single initialization
sentence changed collective payoff more than three times as much as the
full range of the topology effect. The Bayesian agents are task-specific baselines rather than
general upper bounds, and the informative comparison with the LLM
agents is behavioral (Section~\ref{sec:behavior}).

We note several limitations of this work. First, all LLM agents
use gpt-oss-120b with one prompt template and medium reasoning
effort. The center-start and copying patterns may therefore be
specific to this configuration.
Second, since the original Mason--Watts landscapes differ from ours and are not publicly available, human comparisons are limited to the direction of the topology association and to behavioral measures that do not depend on landscape identity. Third, agents act synchronously, and
asynchronous protocols may affect convergence rates and the
topology effect. Fourth, the $15$-round horizon is retained for
comparability with the original experiment. Topology mainly affects
information-diffusion speed, which is reflected in cumulative but
not final-round payoff. Longer horizons would give the converged
phase more weight and reduce the measured cumulative effect. Fifth,
the two-dimensional spatial search task was chosen for comparability
with the human benchmark. Transfer to richer tasks such as planning
or coding is not tested. These limitations motivate tests with additional LLMs and
prompting configurations, alternative network and update protocols,
and richer collaborative tasks.

%\input{ack.tex}

% \clearpage

% Reducing font size (to 9pt) for References & Author Biagraphies
%\footnotesize

%\bibliographystyle{IEEETran}
\bibliographystyle{splncs04}
% AUTHOR: Include your bib file here
\bibliography{refs.bib}

\end{document}